\documentclass[a4paper,conference]{IEEEtran}
\usepackage{cite}
\usepackage{adjustbox}
\usepackage{amsmath}
\usepackage{comment}
\usepackage{bm}
\usepackage[normalem]{ulem}
\useunder{\uline}{\ul}{}
\usepackage{algorithm} 
\usepackage{algpseudocode} 
\usepackage{multirow}
 \usepackage{cuted}
\ifCLASSINFOpdf
\else
\fi

\begin{document}

 This paper is a preprint (Accepted in ICPR2022).\\

\textcopyright  2022 IEEE. Personal use of this material is permitted. Permission from IEEE must be obtained for all other uses, in any current or future media, including reprinting/republishing this material for advertising or promotional purposes, creating new collective works, for resale or redistribution to servers or lists, or reuse of any copyrighted component of this work in other works.

\clearpage
%
\title{CES-KD: Curriculum-based Expert Selection for Guided Knowledge Distillation}

\author{\IEEEauthorblockN{Ibtihel Amara}
\IEEEauthorblockA{McGill University\\
Montreal, Canada\\
\small ibtihel.amara@mail.mcgill.ca}
\and
\IEEEauthorblockN{Maryam Ziaeefard}
\IEEEauthorblockA{McGill University\\
Montreal, Canada\\
\small maryam.ziaeefard@mcgill.ca}
\and
\IEEEauthorblockN{Brett H. Meyer}
\IEEEauthorblockA{McGill University\\
Montreal, Canada\\
\small brett.meyer@mcgill.ca}
\and
\IEEEauthorblockN{Warren Gross}
\IEEEauthorblockA{McGill University\\
Montreal, Canada\\
\small warren.gross@mcgill.ca}
\and
\centering
\IEEEauthorblockN{James J. Clark}
\IEEEauthorblockA{McGill University \\
Montreal, Canada \\
\small james.j.clark@mcgill.ca}
}

%


\maketitle

\begin{abstract}
Knowledge distillation (KD) is an effective tool for compressing deep classification models for edge devices. However, the performance of KD is affected by the large capacity gap between the teacher and student networks. Recent methods have resorted to a multiple teacher assistant (TA) setting for KD, which sequentially decreases the size of the teacher model to relatively bridge the  size gap between these models.  
This paper proposes a new technique called \emph{Curriculum Expert Selection for Knowledge Distillation (CES-KD)} to efficiently enhance the learning of a compact student under the \emph{capacity gap problem}. This technique is built upon the hypothesis that a student network should be guided gradually using stratified teaching curriculum as it learns easy (hard) data samples better and faster from a lower (higher) capacity teacher network.
Specifically, our method is a gradual TA-based KD technique that selects a single teacher per input image based on a curriculum driven by the difficulty in classifying the image. 
In this work, we empirically verify our hypothesis and rigorously experiment with CIFAR-10, CIFAR-100, CINIC-10, and ImageNet datasets and show improved accuracy on VGG-like models, ResNets, and WideResNets architectures.   
\end{abstract}

%
\IEEEpeerreviewmaketitle
\section{Introduction}
Modern deep networks are over-parameterized and are computationally expensive to be deployed onto edge devices. There have been a tremendous focus on compressing large models in the literature \cite{han2015learning,guo2016dynamic,wu2016quantized,denton2014exploiting,luo2016face,zoph2018learning}. Knowledge distillation (KD) \cite{hinton2015distilling}, a model compression technique, which relies on a teacher-student training protocol, has gained popularity in the past few years. The success of KD is mainly associated to its noticeable versatility and generalization aspect. In other words there are no constraints on the type of network architecture. Instead, "any teacher model can teach any student" \cite{cho2019efficacy}, to a certain extent.

However, \emph{"every success hides in it some multiple shortcomings"}. In point of fact, it has been shown in \cite{TAKD} that performing KD does not always yield better student performance. KD might not succeed if the capacity of the student network is much lower than the teacher's capacity. This phenomenon is called the \emph{capacity gap problem} in KD. 
Multiple teacher assistants (TA) of intermediate capacity sizes has become a go-to technique to overcome this shortcoming \cite{TAKD, DGKD}. However the sequential distillation process in \cite{TAKD} and the densely guided ensemble distillation process for learning TAs \cite{DGKD} might not help a lot in enhancing the compressed student's performance since the former depends solely on a single TA network and the latter exploits the average (i.e. aggregated) knowledge of the TAs and ignores the diversity and importance of each expert (i.e. TA networks) within the ensemble.
\begin{figure}[t]
\begin{center}
\includegraphics[width=1\linewidth]{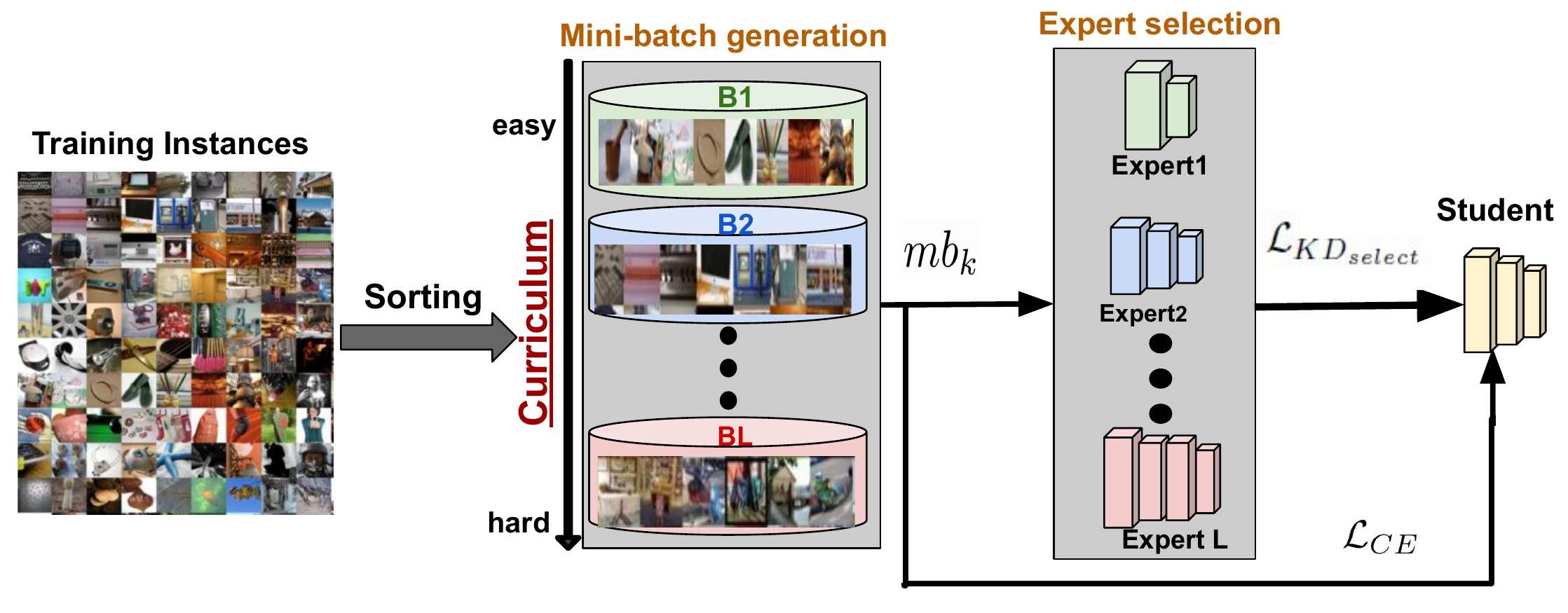}
\caption{\textbf{Overall Pipeline of the proposed \emph{CES-KD} framework.} Given a training dataset, we design an easy-to-hard curriculum on the data using a meta-network (described in \ref{sec: Methodology}). Then, we bucketize the sorted training set according to the total number of available teacher and assistant networks (i.e. experts). The student network is guided via distillation by selecting a single expert within the pool of teacher and assistant models. This expert selection is determined by $\mathcal{L}_{{KD}_{select}}$ loss between the expert and the student. The total loss of CES-KD pipeline consists of both the distillation loss $\mathcal{L}_{{KD}_{select}}$ and the cross entropy loss between the student and the sorted data $L_{CE}$}
\label{fig:ceskd_overview}
\end{center}
\end{figure}

Intuitively, the learning process of a compressed student network can benefit from a stratified teaching and sub-curriculum oriented KD. Indeed, the student's training behaviour changes when faced with easy-to-hard curriculum on the data \cite{weinshall2018curriculum} and also when faced with different teacher or experts in terms of size for distillation.
Therefore, we hypothesize that a compact student network has faster and better learning ability on easy data samples when guided by a less complex or lower capacity teacher network. Similarly, a compact student learns better and faster on difficult data samples when guided by a more complex or higher capacity teacher model during training.

We propose \emph{Curriculum Expert Selection for Knowledge Distillation (CES-KD)}, a TA-based KD technique that is established on a single expert selection per input sample to guide the KD process from a large cumbersome teacher network down to a compressed low capacity student network. 
Our method borrows insights from the field of curriculum learning and adopts the multiple teaching assistant scheme \cite{TAKD,DGKD}, but mainly leverages the individuality of each TA network within the ensemble. In particular, our curriculum learning approach can be globally summarized as follows: given the level of classification difficulty of an input image, we assign an appropriate expert (i.e teacher network) for distillation. More details can be found in Section \ref{sec: CESKD}.
The main contributions of our paper are: 
\begin{enumerate}
\item We propose a curriculum based KD approach that intuitively guides the learning process of a compact student network by selecting the appropriate teacher assistant network according to the provided input samples.

\item We empirically show that on easy data samples, the compressed student learns better and faster from low capacity teachers and on difficult data samples the student learns better and faster from higher capacity teachers.

\item We improve the accuracy of the student network on various datasets and architectures as compared to baseline and state of the art methods.

\item We show that our method has faster convergence than state of the art methods.  

\end{enumerate}
\begin{figure*}[t]
\begin{center}
\includegraphics[width=\linewidth,height=7cm]{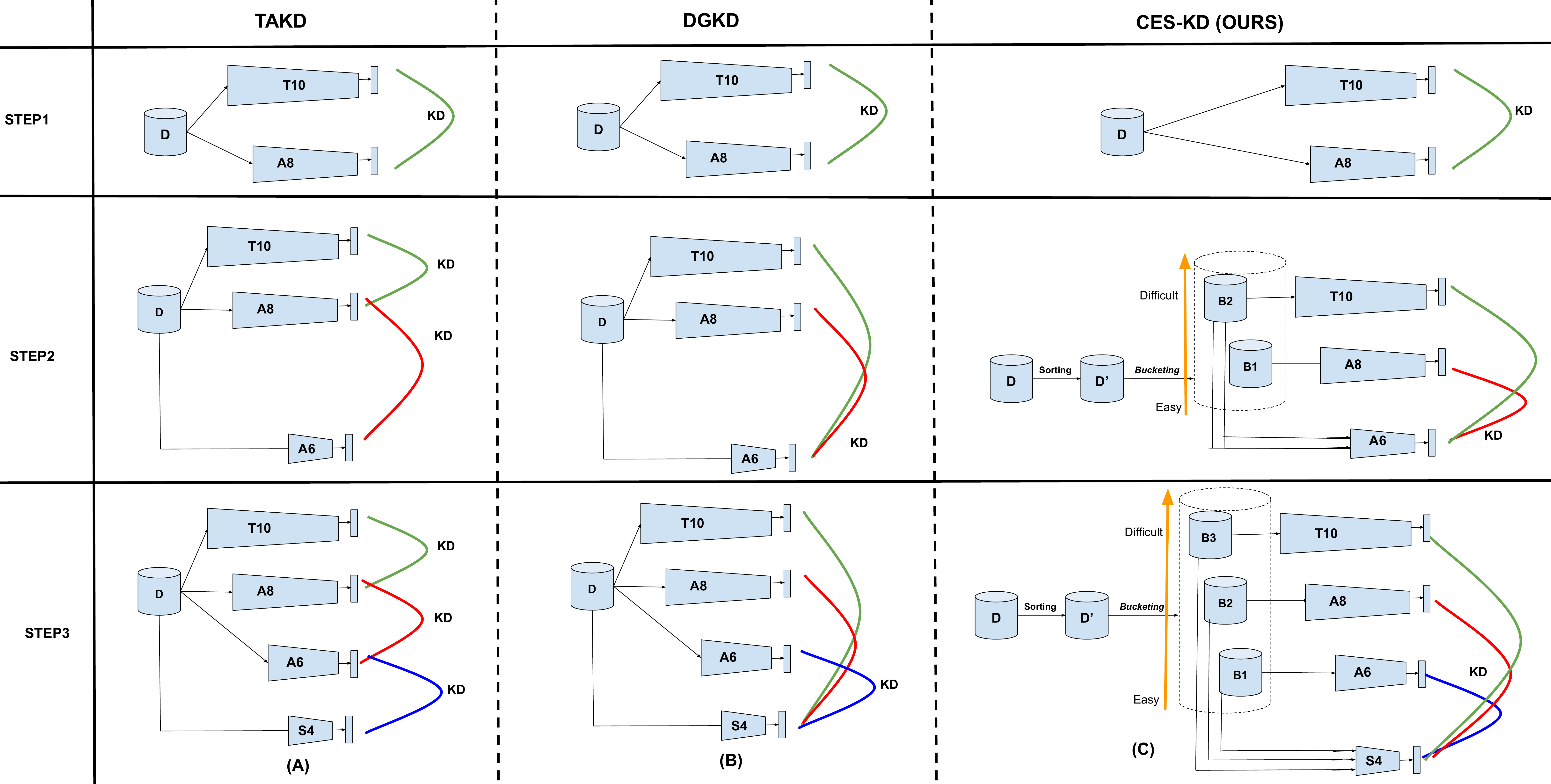}
\caption{\textbf{Back-to-back comparison of the distillation process of TAKD (first column), DGKD (middle column) and CES-KD (last column) methods.} Three-step distillation process from a teacher network of 10 layers (T10) down to a student network of 4 layers (T4), with intermediate teacher assistants A8 and A6. All three methods share the first distillation step giving the assistant network A8. As for the second distillation, TAKD performs a baseline KD from A8 to A6. DGKD performs ensemble KD of T10 and A8 to A6. CES-KD (our method) is a two-step process. First, we sort the data samples by level of difficulty using a scoring function (This process is fixed and is done only once during the entire distillation processes). Second, we bucket the sorted training dataset (equal division of samples into buckets) and assign each bucket to a designated expert (i.e. teacher / assistant network). The bucket containing the easiest data samples is given to the latest teacher assistant and the bucket containing the most difficult samples in the dataset are set to the large teacher network. This bucketing technique is dynamic: at each distillation step the sorted data is equally divided to the total number of experts. (For example two assistants and one teacher network yields to three buckets in total.) }
\label{fig:ceskd}
\end{center}
\vspace{-0.5cm}
\end{figure*}

\section{Related works}
\subsection{Knowledge Distillation}
Knowledge distillation is a training method that is based on teacher-student learning. It was firstly introduced as a mode of model compression by Bucilua et al. \cite{bucilua2006model} then further popularized by Hinton et al. \cite{hinton2015distilling}. The goal of KD is to increase the accuracy of a student network due to transfer of information from the pre-trained teacher network during training. 
Distillation methods rely on different techniques to capture the knowledge of the teacher that is transferred to the student. The traditional KD \cite{hinton2015distilling} uses the soft targets produced by the teacher network as the knowledge to transfer to the student. Some recent works focus on feature-based knowledge distillation and train the student to match the intermediate layers of the teacher \cite{romero2014fitnets}. In addition, some approaches \cite{tung2019similarity} transfer spatial attention maps, where the student attends to similar parts of the image as the teacher network. In our work we exploit the soft targets of the teacher network for our distillation process. 

\subsection{Capacity gap problem}
It was empirically shown in \cite{TAKD} that compressed student networks are harder to train when the size gap between the teacher and the student model is very large. 
Many papers have attempted to solve this issue \cite{TAKD,AKD,DGKD}.
Notably KD techniques that used multiple teacher assistant (TA) networks have stood out to relatively solve the capacity gap problem in KD.
TAKD\cite{TAKD} proposed using intermediate capacity networks to distill Knowledge from a large dense network down to a compact student. DGKD \cite{DGKD} is a method to mitigate the capacity gap problem through densely performing the TAKD process for knowledge distillation. In this way each teacher assistant will learn from the ensemble of previous teacher assistants and teachers. Similarly, the student will learn from all predefined teacher assistants and the main teacher.
Our method differs from these techniques as we use a curriculum-based paradigm to train the student network, using the generated multiple assistant models from the large teacher network. We explicitly utilize the obtained teaching assistant models and the large teacher model to guide the learning of the student. Unlike the work in \cite{DGKD}, we do not aggregate the knowledge of these multiple assistants and teacher to train the student model. Instead, we exploit the diversity of these networks and their individual expertise on the training data through the expert selection aspect of our method. 

\subsection{Curriculum learning in KD}
 The field of curriculum learning (CL) became popular in deep learning due to its ability to alleviate certain training problems by tackling the structure of the training data instead of the network architecture. CL imposes an order on the training process of a student network. It was shown that this technique efficiently enhances the performance of deep models \cite{bengio2009curriculum}.
 The CL process consists of two major steps: (1) a scoring function, which organizes the data by level of difficulty and (2) a pacing function, which defines the process of feeding the sorted data into the network. 
 There are various methods to score and sort a dataset and to perform pacing functions. These were investigated by Hacohen and Weinshall \cite{hacohen2019power}.
 CL was also shown to be beneficial for KD.  Xiang et al. \cite{xiang2020learning} trained a student network and implemented a self-paced learning function to classify imbalanced datasets and showed a substantial boost in performance. Zhao et al. \cite{zhao2021knowledge} performs an instance-level KD where the curriculum is applied on the data and is ensured by a snapshot copy of a student network.  
 Panagiotatos et al. \cite{curriculum_model} applied curriculum on the teacher network rather than on the data. They showed that teachers of different learning levels can guide the compact student's training during distillation. In their work, they took different versions of a teacher network at different training point and used them to perform ensemble knowledge distillation on the entire dataset. 
In our work, we focus on applying curriculum on both the dataset and the teacher models. We sort the dataset based on an easy-to-hard curriculum. Then, we perform a bucketing mechanism, which divides the sorted dataset into different buckets according to the available teacher and teacher assistants. Each bucket is assigned to a specialized teacher assistant or teacher network based on its level of expertise. More details can be found in Section \ref{sec: CESKD}. 

\section{Curriculum-based Expert Selection for KD}
\label{sec: CESKD}
\subsection{Hypothesis and Motivation}
\begin{figure}[h]
    \begin{minipage}{1\linewidth}
    \begin{minipage}{0.49\textwidth}
    \includegraphics[width=1\linewidth]{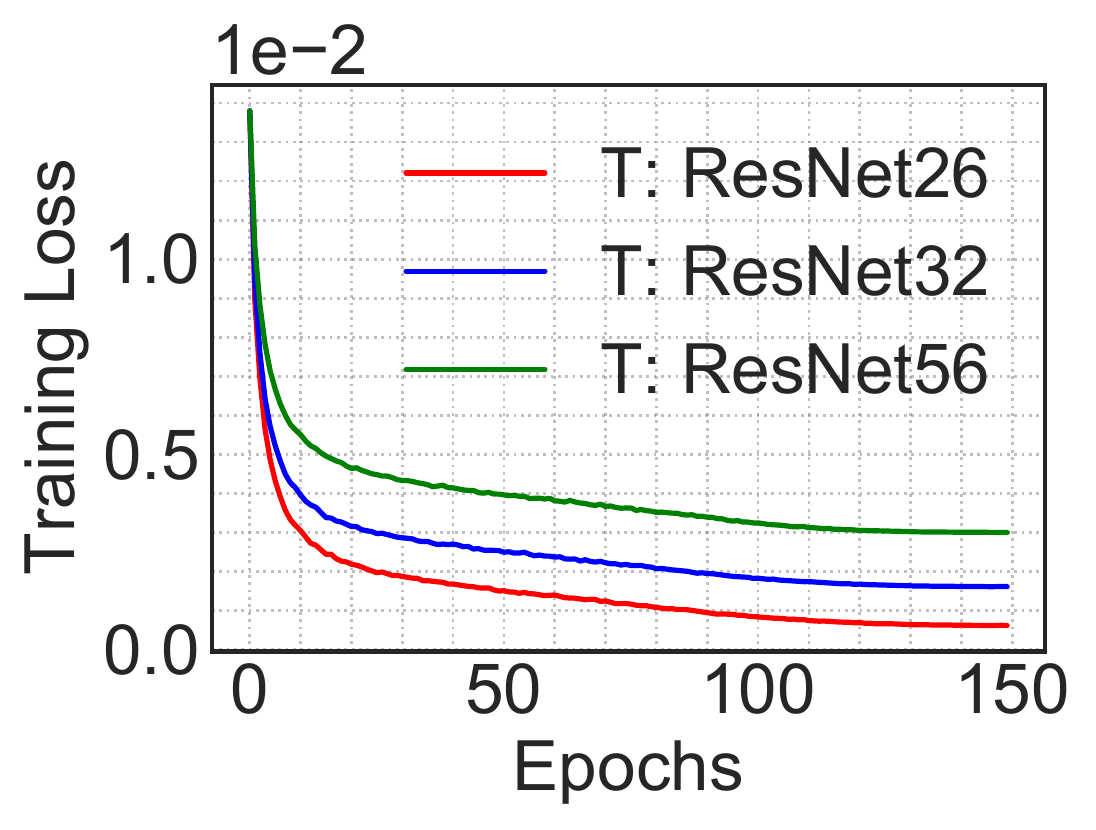}
    \centering{(a)}
    \end{minipage}
    \begin{minipage}{0.49\textwidth}
    \includegraphics[width=1\linewidth]{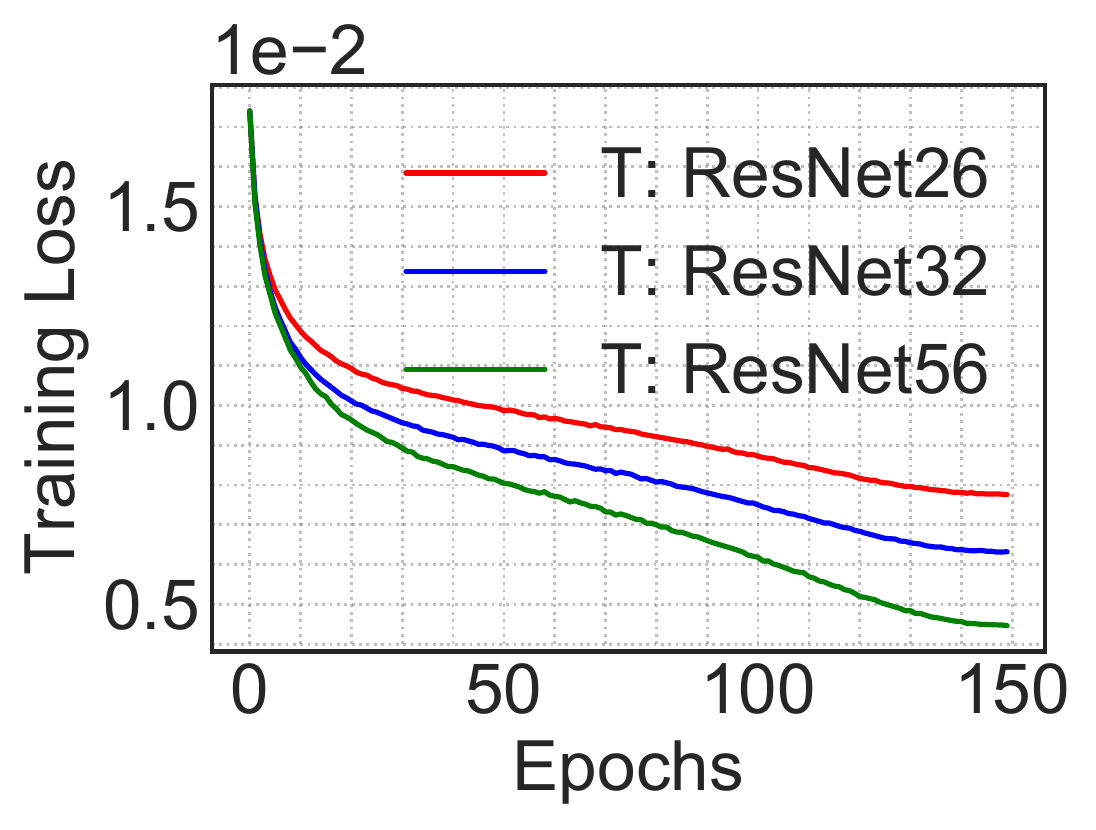}
    \centering{(b)}
    \end{minipage}
    \hfill
    \caption{Training loss of a compact student network (S:ResNet20) distilled using baseline knowledge distillation technique (BLKD) on easy data samples (a) and on hard data samples (b) from different teacher networks of different capacities (T: ResNet26, ResNet32, ResNet56).}
    \label{fig:hypothesis}
    \end{minipage}
\end{figure}

We hypothesize that a student network learns better and faster from a small TA network when faced with easy concepts and better and faster from a large teacher network when dealing with difficult concepts.
To validate this hypothesis, we trained a compact student network (ResNet20) on CINIC-10 dataset. The training data are sorted according to an easy-to-hard curriculum using a finetuned meta-network (full details on designing this curriculum is given in Section \ref{subsub:scoringfn}. We then divide the obtained curriculum into three separate subsets of different levels of difficulty: Easy, Intermediate and Difficult. We also took different teacher networks of different sizes (i.e capacities) to guide the student's training via baseline knowledge distillation (BLKD)\cite{hinton2015distilling} on these different levels of curriculum.
Table \ref{tab:ablation1} shows the test accuracy of the student network trained on three different levels of difficulty with three teachers of different capacities. We see that for easy samples the student network (ResNet20) has higher accuracy when guided through distillation by the lowest capacity teacher (ResNet26) within the group of experts. As for the difficult samples, the student acquires better knowledge from the highest capacity teacher model (ResNet56), which validates our assumption in terms of quality of learning.
To further study the student's learning efficiency, Figure \ref{fig:hypothesis} (a) shows that the student's optimization, on easy samples, converges faster when trained with the guidance of the lowest capacity teacher model (ResNet26) via knowledge distillation. Also in Figure \ref{fig:hypothesis} (b) we observe that on difficult samples, the student learns faster (i.e. faster convergence) from the highest capacity network (ResNet56) than from the lower capacity teachers. This comparison validates our hypothesis and motivates our technique regarding the curriculum data-model selection.

\begin{table}[h]
\centering
\caption{Top-1 \% test accuracy of the student network (ResNet20) trained on three curriculum levels under the supervision (distillation) of different capacity teacher networks on the test set of CINIC10 dataset. Average over three independent runs. }
\label{tab:ablation1}
\resizebox{\linewidth}{!}{%
\begin{tabular}{|c|c|c|c|}
\hline
\textbf{Teacher networks} & \textbf{Easy} & \textbf{Intermediate} & \textbf{Difficult} \\ \hline
\textbf{Resnet 56} & 79.08 \% & 81.59 \% & \textbf{79.20\%} \\ \hline
\textbf{Resnet 32} & 79.28 \% & \textbf{81.88 \%} & 79.10 \% \\ \hline
\textbf{Resnet 26} & \textbf{79.30 \%} & \textbf{81.88 \%} & 78.97 \% \\ \hline
\end{tabular}
}
\end{table}

\label{sec: Methodology}
\subsection{Methodology of our proposed method}
Figure \ref{fig:ceskd_overview} presents a global overview of the distillation framework CES-KD. 
Our method relies on two main steps: the design of the data curriculum and the selection of a single representative expert based on their expertise on a given input data (i.e. images).
We provide in Figure \ref{fig:ceskd} details on the distillation pipeline of our method with a back-to-back comparison on current TA-based KD methods (TAKD \cite{TAKD} and DGKD\cite{DGKD}).
Following the standard CL paradigm, we need to address two main questions: (1) How do we rank the dataset according to an easy-to-hard curriculum? (see subsection \ref{subsub:scoringfn}); (2) How do we train a student model using the ranked instances? (See subsection \ref{subsub:rts}).
\subsubsection{Design of the data curriculum}
\label{subsub:scoringfn}
In this work, we adopt the \emph{meta-network} method described by Hacohen and Weinshall \cite{hacohen2019power} as \emph{transfer learning-based scoring function}. In particular, we consider a reference model trained on a very large dataset. Then, we fine-tune this reference model on the smaller training dataset. 
We evaluate the difficulty score of the training data using the real-valued loss given by the fine-tuned meta-network. This method was also previously explored in \cite{behnam_curricula}.
Therefore, for a given reference model with weights $\bm{W}$,  $f_{\bm{W}} : \bm{X} \xrightarrow{} \bm{Y}$, the difficulty score of a sample $x_i$ is defined by $s(x_i,y_i) = Loss(f_{\bm{W}}(x_i),y_i)$. Finally, these samples are sorted in an ascending order based on their score values, resulting in a sorted dataset called $D_{sorted}$. This curriculum specification is performed only once during our entire distillation pipeline. 
\subsubsection{Representative expert selection}
\label{subsub:rts}
After ranking the instances within the training dataset, we perform a \textit{bucketing system} in which we split $D_{sorted}$ into $L$ equal buckets $\{B_1,...,B_L\}$  according to the number of experts (teacher and teacher assistants) we are considering for distillation, at a particular distillation step. The first bucket contains the easiest samples and the last bucket includes the hardest samples. Each bucket is assigned to a representative teacher or TA network based on their level of expertise. The first bucket containing the easiest samples are given to the lowest capacity teacher network. The next bucket is given to the second lowest capacity teacher assistant network, until the last bucket containing the hardest data samples are sent to the largest teacher network. 

\begin{algorithm}
    \caption{CES-KD}

    \begin{algorithmic}[1]
    
\State Rank the original dataset $D$ according to  scoring function $s$: $D_{ranked}$

\State Divide $D_{ranked}$ into $L$  equal buckets: {$\{B_1,...,B_L\}$}.
       
       \State Assign the experts in an ascending order of depths to the buckets, i.e the shallowest expert is assigned to $B_1$ until the deepest expert is assigned to $B_L$
       
       \State Generate mini-batches $mb_k$ in each bucket
           \For {all mini-batches in $D_{ranked}$}
              \For {each bucket $B_l$ in [$B_1$,...,$B_L$]}
               \If {$mb_k$ belongs to $B_l$}
                \State {Select the representative expert $l$ from the ensemble to provide soft targets for $mb_k$} 
                \EndIf
            \EndFor
            \State Update student weights with $mb_k$ 
        \EndFor
        
    \end{algorithmic}

\end{algorithm}

\subsubsection{Distillation loss}
\label{subsub:distillation}
Following the proposed data curriculum and teacher selection, we train the student network according to the ascending level of difficulty of instances in $D_{ranked}$. We generate mini-batches $mb_k$ within each bucket $B_l$, where $k$ is the number of mini-batches per bucket, and guide the student with the output logits produced by the selected expert. $l$ is the index of the representative teacher or TA network. We define the distillation loss as: 
\begin{equation}
   \mathcal{L}_{{KD}_{select}} = \sum_{l=1}^{L}{\omega_l^{mb_k \in B_l} \mathcal{L}_{CE}(\sigma{(\frac{ \bf{z}_s}{T}}), \sigma(\frac{\bf{z}_l}{T}) ) }
\end{equation}
where: 
\begin{equation*}
    \omega_l^{mb_k \in B_l} = \bm{1}_{B_{l}}(mb_i) =
    \begin{cases}
      1 & \text{if $mb_k \in B_l$}\\
      0 & \text{else}\\
    \end{cases}       
\end{equation*}
$\mathcal{L}_{CE}(.,.)$ denotes the cross entropy loss.
$\sigma(.)$ is the softmax function. $\bf{z}_s$ and $\bf{z}_l$ are output logits of the student and selected expert $l$, respectively. $T$ is the temperature hyperparameter.
$\omega_l^{mb_k \in B_l}$ denotes a weight that selects the representative experts from the ensemble according to the input mini-batch. 

\subsubsection{Training}
The total training loss is then defined as: 
\begin{equation}
    \mathcal{L_{CES-KD}} = \alpha T^2 \mathcal{L}_{KD} + (1-\alpha) \mathcal{L}_{CE}(\hat{\bf{y}},\sigma(\bf{z}_s))
\end{equation}
$\hat{\bf{y}}$ is the one-hot vector indicating the ground-truth class. $\bf{z_s}$ is the output logit vector of the student model. The pseudo code of our overall methodology is provided in Algorithm 1.
\section{Experimental Set-up}
\label{sec:experimental_setup}
\subsection{Datasets and Networks}
We perform experiments on different datasets - CIFAR-10 \cite{krizhevsky2009learning}, CINIC-10 \cite{darlow2018cinic}, CIFAR-100 \cite{krizhevsky2009learning}, and ImageNet\cite{deng2009imagenet}, having different numbers of classes ranging from 10, 100 up to 1000. We also used different network architectures to validate the performance of our method from plain convolutions (plain CNNs), which are VGG-like networks \cite{simonyan2014very}, ResNets \cite{he2016deep}, and WideResNets \cite{zagoruyko2016wide}. We report the performance in terms of accuracy on the validation set for ImageNet and on the test set for the remaining of the datasets. 

\subsection{Implementation Details}
All implementations are done using PyTorch\cite{pytorch}. For all experiments we perform data augmentation, specifically random crops and random horizontal flips. Then we perform data normalization by subtracting the mean and the variance of the entire training set. For the experiments on plain CNN architectures on CIFAR-100 and ResNets on CIFAR-10, we used stochastic gradient descent (SGD) optimizer with Nesterov momentum of 0.9, weight decay of 1e-4, and with a mini-batch size of 128. The initial learning rate was 0.1, then divided by 10 at the 30th, 90th, and 120th epochs; we trained for a total of 150 epochs. Unlike TAKD \cite{TAKD} and DGKD \cite{DGKD}, we do not use a hyper-parameter optimization toolkit on a hyper-parameter search space and seed setting. Instead, we perform multiple runs of the experiment and report the average across multiple random seeds, and the corresponding standard deviations. For fairer comparison between techniques we used the same hyper-parameters for each distillation step and across different TA-based KD techniques. As for KD related hyper-parameters, we took $\alpha$ to be 0.9 and the temperature $T$ to be 10.
For the experiments related to benchmarking on CIFAR-100 and comparing our method to state-of-the-art KD techniques, we use the same set-up as Tian et al. in \cite{tian2019crd} detailed in their code\footnote[1]{https://github.com/HobbitLong/RepDistiller}. For the ImageNet experiments, we use the same training set-up as the Imagenet distributed training from Pytorch \footnote[2]{https://github.com/pytorch/examples/tree/master/imagenet}. 

\subsection{Data Sorting and Curriculum}
For experiments on CIFAR-10, CINIC-10, and CIFAR-100, we use a reference model of ResNet110 pre-trained on ImageNet. The ResNet model was taken from the pretrained torch model zoo in Pytorch \cite{pytorch}. We freeze the feature layers and fine-tune using only the classification layers on the targeted smaller training dataset.
For the experiments related to ImageNet, we do not perform fine-tuning. Instead we consider the pre-trained network itself from \cite{pytorch} as the reference model. We ran extensive trials using the ranked data instances to train the student. We report the results of the curriculum that works best in training our students. All buckets are incrementally presented per epoch to the student network to avoid the issue of catastrophic forgetting \cite{kirkpatrick2017overcoming,kemker2018measuring}. The classes are distributed almost equally across the buckets to avoid any biases during training. We generate mini-batches per bucket with uniform sampling of instances within the bucket. 

\section{Results and Discussion}
\subsection{Comparing CES-KD to current TA-based KD methods}
In this section we show the effectiveness of our proposed method CES-KD on several standard datasets such as CIFAR-10 and CIFAR-100 when compared to TA-based KD methods. 

\begin{table}[!h]
\centering
\caption{Test accuracy with all distillation steps using plain CNN architecture on CIFAR100 over three random seeds. We also report the corresponding standard deviation on the steps containing TA networks. Teacher $T_{10}$ Assistants $A_{8}, A_{6}$  and Student $S_4$. (*) methods use publicly provided code.} 
\resizebox{\linewidth}{!}{ 
\begin{tabular}{|c|c|c|c|}
\hline
\textbf{Step} & \textbf{TAKD*} & \textbf{DGKD*} & \textbf{CES-KD} \\ \hline
Teacher (CNN-10;$T_{10}$) & \multicolumn{3}{c|}{66.89} \\ 
Student (CNN-4; $S_{4}$) & \multicolumn{3}{c|}{62.71} \\ \hline
$T_{10} \rightarrow A_{8}$ & \multicolumn{3}{c|}{64.20} \\ \hline
$T_{10} \rightarrow A_{8} \rightarrow A_{6}$ & $67.80\pm0.35$ & $68.56\pm0.282$& \bm{$ 68.90\pm0.196$} \\ \hline
$T_{10} \rightarrow A_{8} \rightarrow A_{6} \rightarrow S_{4}$ &$63.31\pm0.145$& $ 63.44\pm0.127$& \bm{$ 63.58\pm0.045$}    \\ \hline
\end{tabular}
} 
\label{tab:cnn_cifr100}
\end{table}

Table \ref{tab:cnn_cifr100} shows the test accuracy of a student network (4-layered CNN) when distilled from a teacher network (10-layered CNN) at each level of a defined distillation path. To distill knowledge from a teacher $T_{10}$ down to a student $S_{4}$, we performed the following distillation path $T_{10} \rightarrow A_8 \rightarrow A_6 \rightarrow S_4$. 
Our method shows good improvements overall. CES-KD achieves an accuracy of 68.90 \% for the  $T_{10} \rightarrow A_8 \rightarrow A_6 $ path, which is almost 1\% improvement to TAKD, a notable increase compared to DGKD, and especially a 6.19 \% improvement compared to the student network trained individually and from scratch. As for the path $T_{10} \rightarrow A_8 \rightarrow A_6 \rightarrow S_4$, we also see a substantial accuracy increase with our proposed method.

\begin{table}[!h]
\centering
\caption{Test accuracy with all distillation steps using ResNet architecture on CIFAR10 over three random seed. We also report the corresponding standard deviation on the steps containing TA networks. Teacher $T_{26}$ Assistants $A_{20}, A_{14}$  and Student $S_8$. (*) methods use publicly provided code.} 
\resizebox{\linewidth}{!}{ 
\begin{tabular}{|c|c|c|c|}
\hline
\textbf{Step} & \textbf{TAKD*} & \textbf{DGKD*} & \textbf{CES-KD} \\ \hline
Teacher (ResNet26;$T_{26}$) & \multicolumn{3}{c|}{91.73} \\ 
Student (ResNet8; $S_{8}$) & \multicolumn{3}{c|}{85.35} \\ \hline
$T_{26} \rightarrow A_{20}$ & \multicolumn{3}{c|}{91.44} \\ \hline
$T_{26} \rightarrow A_{20} \rightarrow A_{14}$ & $90.45 \pm 0.122 $& $ 90.66\pm 0.120 $& \bm{$90.81\pm 0.124 $} \\ \hline
$T_{26} \rightarrow A_{20} \rightarrow A_{14} \rightarrow S_{8}$ &$ 86.71 \pm 0.163 $& $86.85 \pm 0.250 $& \bm{$ 86.97 \pm 0.230 $}    \\ \hline
\end{tabular}
} 
\label{tab:resnet_cifr10}
\end{table}

 Table \ref{tab:resnet_cifr10} shows the test accuracy of a compact student network using a residual architecture. The teacher network is a ResNet26 and the student is a ResNet8. Similarly to our previous observations, our method shows a considerable improvement when compared to both TAKD and DGKD. For both paths $T_{26} \rightarrow A_{20} \rightarrow A_{14} $, $T_{26} \rightarrow A_{20} \rightarrow A_{14} \rightarrow A_{8}$, CES-KD shows, respectively, a 5.46\% and a 1.62\% improvement over the student network trained from scratch. 

\subsection{Teacher Selection Ablation}
In this section, we investigate the effect of different method of teacher selection. For this we distilled a compact student network (4-layered CNN) from a teacher (10-layered CNN) on CINIC-10 dataset. We performed three types of teacher selection: (1) baseline selection in which we perform our proposed CES-KD method. Mainly, the bucketed sorted data are assigned as following:  the easiest samples to the lowest capacity teacher assistant and the hardest examples to the largest teacher network; (2) Anti-selection in which we assign the easy buckets to the largest teacher network and the buckets containing the hardest examples are assigned to the lowest capacity teacher assistant network. Finally, (3) random curriculum in which we randomly assign teacher assistants and teachers to the buckets during training. 
\begin{table}[h]
\caption{Test accuracy using Plain CNN architecture on CINIC-10. We considered the distillation path $T_{10} \xrightarrow{} T_{8} \xrightarrow{} T_6 \xrightarrow{} T_4$. We performed different scenarios of teacher selection: Baseline, Anti, and Random selection.}
\label{tab:ablation_curriculum}
\centering
\resizebox{\linewidth }{!}{
\begin{tabular}{|l|l|l|l|}
\hline
\multicolumn{1}{|c|}{\textbf{Selection}} &
  \multicolumn{1}{c|}{\textbf{Baseline}} &
  \multicolumn{1}{c|}{\textbf{Anti}} &
  \multicolumn{1}{c|}{\textbf{Random}} \\ \hline
\textbf{Accuracy} &
  \bm{$71.635 \pm 0.179$} &
  $71.002 \pm 0.172 $ &
  $71.156 \pm 0.036$ \\ \hline
\end{tabular}
}
\end{table}

Table \ref{tab:ablation_curriculum} shows the test accuracy on CINIC-10 of a student network (4-layered plain CNN). We demonstrate that having a less specialized expert on an easy concept samples and a very specialized expert on hard samples enhances the performance of a student model. The anti-selection gave the least performance overall. This validates our previous hypothesis that larger networks might not be the optimal teacher when it comes to teaching simple concepts through simple examples. As for the random curriculum where at each distillation step a random expert is chosen, this just shows that there exists a selection rule that can enhance the performance of the distillation process of a student network. 

\begin{table*}[t]
\caption{Test accuracy (\%) of student networks on CIFAR100 on different state-of-the-art distillation methods using different network architectures. (*) are values provided in \cite{tian2019crd} and (**) are implemented values from publicly available code. Average over 5 runs.}
\label{tab:sota_kd}
\resizebox{\textwidth}{!}{
\begin{tabular}{|c|c|c|c|c|c|c|c|c|c|c|c|c|c|c|c|c|c|}
\hline
\textbf{\begin{tabular}[c]{@{}c@{}}Teacher\\ Student\end{tabular}} & \textbf{NOKD (*)} & \textbf{BLKD (*)} & \textbf{FitNet (*)} & \textbf{AT (*)} & \textbf{SP (*)} & \textbf{CC (*)} & \textbf{VID (*)} & \textbf{RKD (*)} & \textbf{PKT (*)} & \textbf{AB (*)} & \textbf{FT (*)} & \textbf{FSP (*)} & \textbf{NST (*)} & \textbf{CRD (*)} & \textbf{TAKD (**)} & \textbf{DGKD (**)} & \textbf{\begin{tabular}[c]{@{}c@{}}CES-KD\\ (ours)\end{tabular}} \\ \hline
\textbf{\begin{tabular}[c]{@{}c@{}}wrn-40-2\\ wrn-16-2\end{tabular}} & \begin{tabular}[c]{@{}c@{}}75.61\\ 73.26\end{tabular} & 74.92 & 73.58 & 74.08 & 73.83 & 73.56 & 74.11 & 73.35 & 74.54 & 72.50 & 73.25 & 72.91 & 73.68 & {\ul 75.48} & 75.25 & {\ul {\ul 75.38}} & \textbf{75.70} \\ \hline
\textbf{\begin{tabular}[c]{@{}c@{}}ResNet110\\ ResNet20\end{tabular}} & \begin{tabular}[c]{@{}c@{}}74.31\\ 69.06\end{tabular} & 70.67 & 68.99 & 70.22 & 70.04 & 69.48 & 70.16 & 69.25 & 70.25 & 69.53 & 70.22 & 70.11 & 69.53 & { \ul {\ul  71.46}} &71.15  & { \ul 71.48} &  \textbf{71.59}\\ \hline
\end{tabular}
}
\vspace{-0.2cm}
\end{table*}

\subsection{CES-KD for faster student training}
\begin{figure}[h]
    \begin{minipage}{1\linewidth}
    \begin{minipage}{0.49\textwidth}
    \includegraphics[width=1\linewidth]{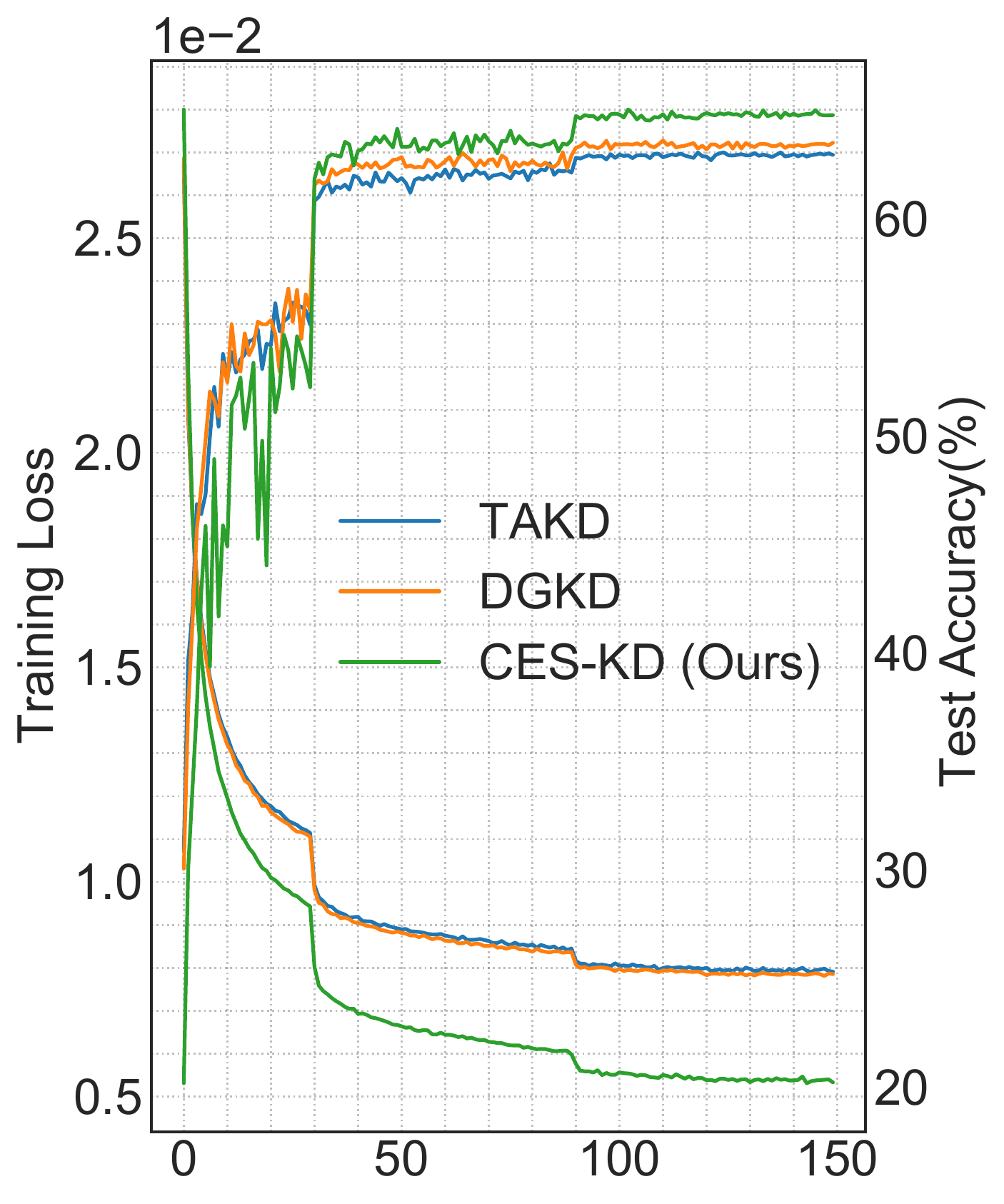}
    \centering{(a)}
    \end{minipage}
    \begin{minipage}{0.49\textwidth}
    \includegraphics[width=1\linewidth]{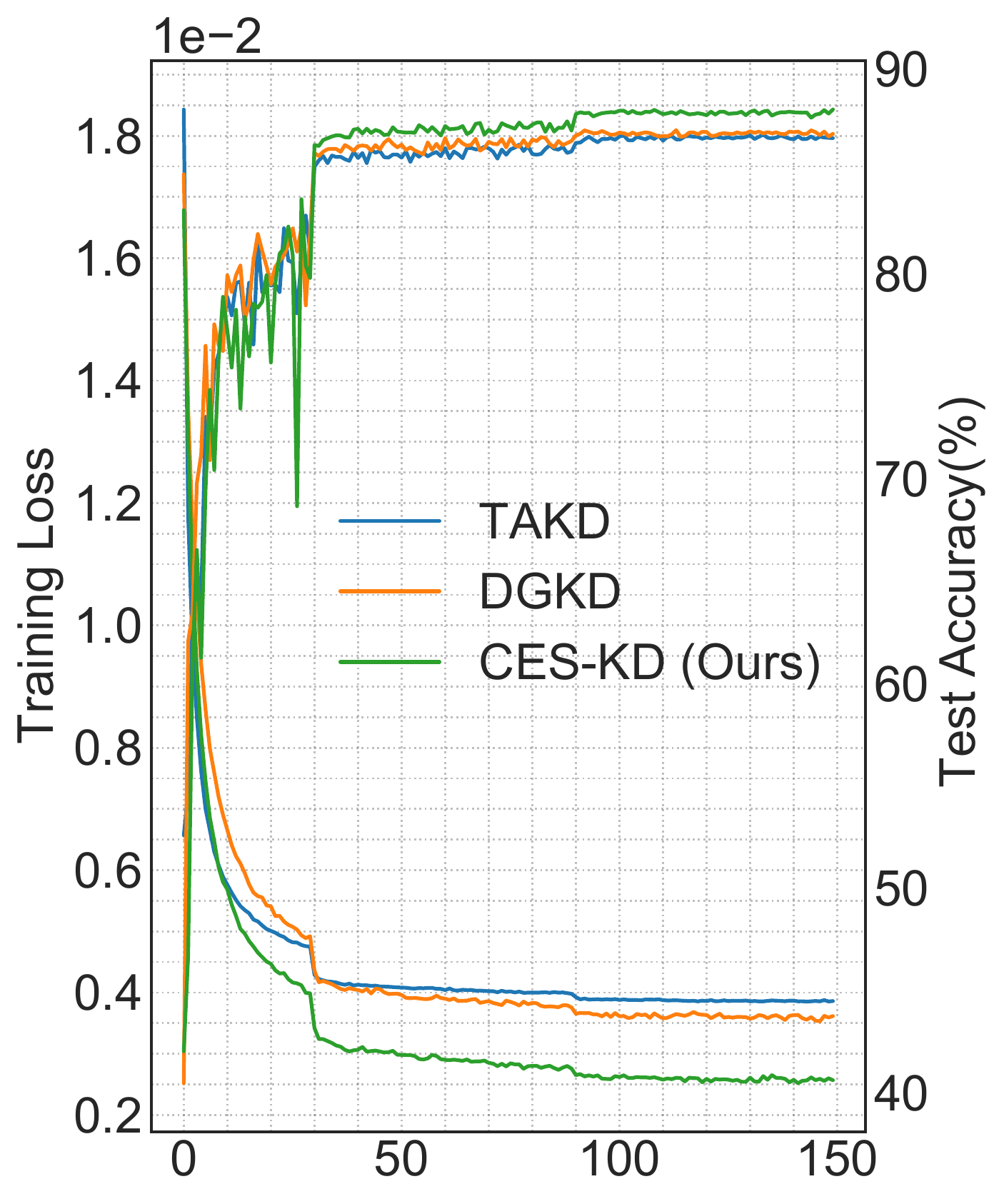}
    \centering{(b)}
    \end{minipage}
    \hfill
    \caption{Training loss and Test Accuracy  of a compact student network 4-layered CNN network on CIFAR-100 test set (a) and ResNet8 on CIFAR-10 test set (b) using different TA-based KD methods TAKD, DGKD, and CES-KD (ours).}
    \label{fig:fast_conv}
    \end{minipage}
    \vspace{-0.2cm}
\end{figure}

TAKD and DGKD methods play a distinctive role in bridging the capacity gap problem. However, for edge devices where the training or fine-tuning time is limited, performing ensemble guidance to the training of the student can be impractical to implement. 
In Figure \ref{fig:fast_conv} we show the training loss and the test accuracy curves over epochs of a 4 layer CNN model and ResNet8 student implemented for all TAKD, DGKD, and CES-KD (ours) method. We see that our method converges faster than DGKD and TAKD. Indeed, we see a fast drop in the training loss with CES-KD using both architectures 4-layered CNNs and ResNet8, after epoch 30 when compared to TAKD and DGKD. Our technique leads to higher performance in fewer epochs. In fact, to reach a target test accuracy of 61\%, our method needs only 30 epochs whereas TAKD and DGKD reach this target value only by epoch 90 for the plain CNN architecture on CIFAR-100. Similarly, for the ResNet8 architecture, if a target test accuracy is set at around 86\%, the CES-KD method reaches this value by epoch 30 while TAKD and DGKD take much more epochs to reach it.
This rapid convergence of the training curve of the student is linked to the curriculum on both data and teacher assignment. This was previously observed in other works in CL \cite{wu2020curricula}. 
\subsection{CES-KD vs SOTAs in KD}
To assess the effectiveness of our method, we compare its test performance to current state-of-the-art KD methods, such as BLKD\cite{hinton2015distilling}, FitNet\cite{romero2014fitnets}, AT\cite{komodakis2017paying}, SP\cite{tung2019similarity}, CC\cite{peng2019correlation}, VID\cite{ahn2019variational}, RKD\cite{park2019relational}, PKT\cite{pathak2016context}, AB\cite{heo2019knowledge}, FT\cite{kim2018paraphrasing}, FSP\cite{yim2017gift}, NST\cite{huang2017like}, CRD\cite{Tian2020ContrastiveRD}, TAKD\cite{TAKD},and DGKD\cite{DGKD}. In Table \ref{tab:sota_kd}, we provide the test accuracies on CIFAR-100 dataset and using two different architectures WideResNets and ResNets. The bold values are the methods that are outperforming, the single underlined values are the second best and the double underlined values show the third best overall. 
For TA-based methods (i.e. TAKD, DGKD, and CES-KD), we adopt the following distillation paths for each architecture network: (1) WRN $40 \times 2 (T) \rightarrow$ WRN $34 \times 2 (A_1) \rightarrow$ WRN $ 22 \times 2 (A_2) \rightarrow$ WRN $ 16\times 2 (S)$, and (2) ResNet110 $\rightarrow$ ResNet56 $\rightarrow$ ResNet44 $\rightarrow$ ResNet32 $\rightarrow$ ResNet20.
We observe globally that our method substantially outperforms some of these KD techniques. Mainly we witness an improvement of 2.44\% from the student network trained individually and solely from data. Similarly, we see the similar trend for the ResNet architecture having a test accuracy on CIFAR-100 of 71.59\%. 

\subsection{Results on ImageNet}
In order to assess the scalability of our method to large datasets, we apply CES-KD to the ImageNet dataset. We chose ResNet56 as our teacher and ResNet18 as our student. The distillation path is ResNet56 (T) $\rightarrow$ ResNet34 ($A_1$) $\rightarrow$ ResNet18 (S).
Table \ref{tab:imagenet} shows top1\% and top5\% validation accuracy on ImageNet. Our method achieved 69.96\% top 1\% accuracy compared to the baseline vanilla KD, BLKD, which reaches 68.94\%. This demonstrates the scalability of our method to larger datasets.
\begin{table}[h]
\renewcommand{\arraystretch}{1.1}
\caption{Top 1 accuracy (\%) on ImageNet. The distillation path is: ResNet34 $\xrightarrow{}$ResNet26 $\xrightarrow{} ResNet18 $ }
\label{tab:imagenet}
\centering
\begin{tabular}{|c|c|c|c|c|c|c|}
\hline
 & Teacher & Student & \begin{tabular}[c]{@{}c@{}}BLKD\\ (KD)\end{tabular} & TAKD & DGKD & CES-KD \\ \hline
Top1\% & 73.1 &  68.54 & 68.94 & 69.03 & 69.77 & \textbf{69.96} \\ \hline
Top5\% & 91.2  & 87.85 &87.88  & 88.21 & 89.52 &  \textbf{89.98}\\ \hline
\end{tabular}
\end{table}

\section{Conclusion}
In this paper, we proposed a curriculum guided expert selection for bridging the capacity gap problem in KD. We follow the TA-based KD method but instead of guiding each TA network sequentially or by performing aggregations, we specifically exploit a curriculum on both data and available TA networks to guide the student's distillation process through a stratified manner of learning. Empirically, we have shown that the learning process of a student through KD is dependent on the difficulty of the samples and also based on the quality of knowledge it is getting from the representative expert. We have demonstrated that a compact student network learns better with lower capacity Teacher/TA networks on easy data samples. Similarly, this compact student's learning benefits from the knowledge  given by larger capacity teacher/TA networks. Our thorough experiments showed that our method can substantially improve the compact student's performance and that it is comparable to current state-of-the art and well-performing  KD techniques. Finally, we also have shown that our method is scalable to larger datasets. This research was enabled in part by support provided by Compute Canada (www.computecanada.ca).
\newpage

\bibliographystyle{./IEEEtran}
\bibliography{./bibliography}

\end{document}